\documentclass[11pt]{article}
\usepackage{paclic34}
\usepackage{times}
\usepackage{latexsym}
\usepackage{amsmath}
\usepackage{multirow}
\usepackage{url}

\usepackage{graphicx}
\usepackage[utf8]{vietnam}
\usepackage{tikz}
\usepackage{pgfplots}
\usetikzlibrary{fit}
\usetikzlibrary{arrows,shapes}
\usetikzlibrary{shapes.misc}

\title{Improving Sequence Tagging for Vietnamese Text using
  Transformer-based Neural Models}

\author{
  The Viet Bui$^{1}$\\
  {\tt vietbt6@fpt.com.vn} \\ \And
  Thi Oanh Tran$^{1,2}$\\ 
  {\tt oanhtt@isvnu.vn} \\ \AND
  Phuong Le-Hong$^{1,2}$\\
  {\tt phuonglh@vnu.edu.vn}\\ \\
  $^1$ FPT Technology Research Institute,
  FPT University, Hanoi, Vietnam\\
  $^2$ Vietnam National University, Hanoi, Vietnam\\
}

\date{}

\usetikzlibrary{shapes.geometric}
\usetikzlibrary{positioning, shadows}

\usetikzlibrary{calc}

\newcommand{\matt}[4]
{
  \node[draw, minimum height=4em, minimum width=3em, 
    fill=purple!30, double copy shadow={shadow xshift=3pt,shadow yshift=3pt, draw}
  ] (#1) at (#2,#3) {#4};

}

\begin{document}

\captionsenglish

\maketitle

\begin{abstract}
  This paper describes our study on using mutilingual BERT embeddings
  and some new neural models for improving sequence tagging tasks for
  the Vietnamese language. We propose new model architectures and
  evaluate them extensively on two named entity recognition datasets
  of VLSP 2016 and VLSP 2018, and on two part-of-speech tagging
  datasets of VLSP 2010 and VLSP 2013. Our proposed models outperform
  existing methods and achieve new state-of-the-art results.  In
  particular, we have pushed the accuracy of part-of-speech
  tagging to 95.40\% on the VLSP 2010 corpus, to 96.77\% on the VLSP
  2013 corpus; and the $F_1$ score of named entity recognition to
  94.07\% on the VLSP 2016 corpus, to 90.31\% on the VLSP 2018
  corpus. Our code and pre-trained models viBERT and vELECTRA are
  released as open source to facilitate adoption and further research.
\end{abstract}

\section{Introduction}
\label{sec:introduction}

Sequence modeling plays a central role in natural language
processing. Many fundamental language processing tasks can be treated
as sequence tagging problems, including part-of-speech tagging and
named-entity recognition. In this paper, we present our study on
adapting and developing the multi-lingual BERT~\cite{Devlin:2019} and
ELECTRA~\cite{Clark:2020} models for improving
Vietnamese part-of-speech tagging (PoS) and named entity recognition
(NER).

Many natural language processing tasks have been shown to be greatly
benefited from large network pre-trained models. In recent years,
these pre-trained models has led to a series of breakthroughs in
language representation learning~\cite{Radford:2018,Peters:2018,Devlin:2019,Yang:2019,Clark:2020}.  Current state-of-the-art
representation learning methods for language can be divided into two
broad approaches, namely \textit{denoising auto-encoders} and
\textit{replaced token detection}.

In the denoising auto-encoder approach, a small subset of tokens of
the unlabelled input sequence, typically 15\%, is selected; these
tokens are masked (e.g., BERT~\cite{Devlin:2019}), or attended (e.g.,
XLNet~\cite{Yang:2019}); and then train the network to recover the
original input. The network is mostly transformer-based models which
learn bidirectional representation. The main disadvantage of these
models is that they often require a substantial compute cost because
only 15\% of the tokens per example is learned while a very large
corpus is usually required for the pre-trained models to be
effective. In the replaced token detection approach, the model learns
to distinguish real input tokens from plausible but synthetically
generated replacements (e.g., ELECTRA~\cite{Clark:2020}) Instead of
masking, this method corrupts the input by replacing some tokens with
samples from a proposal distribution. The network is pre-trained as a
discriminator that predicts for every token whether it is an original
or a replacement. The main advantage of this method is that the model
can learn from all input tokens instead of just the small masked-out
subset. This is therefore much more efficient, requiring less than
$1/4$ of compute cost as compared to RoBERTa~\cite{Liu:2019} and
XLNet~\cite{Yang:2019}. 

Both of the approaches belong to the fine-tuning method in natural
language processing where we first pretrain a model architecture on a
language modeling objective before fine-tuning that same model for a
supervised downstream task. A major advantage of this method is that
few parameters need to be learned from scratch. 

In this paper, we propose some improvements over the recent
transformer-based models to push the state-of-the-arts of two common
sequence labeling tasks for Vietnamese. Our main contributions in this
work are:
\begin{itemize}
\item We propose pre-trained language models for Vietnamese which are
  based on BERT and ELECTRA architectures; the models are trained on large
  corpora of 10GB and 60GB uncompressed Vietnamese text.
\item We propose the fine-tuning methods by using attentional
  recurrent neural networks instead of the original fine-tuning with
  linear layers. This improvement helps improve the accuracy of
  sequence tagging.
\item Our proposed system achieves new state-of-the-art results on all
  the four PoS tagging and NER tasks: achieving 95.04\% of accuracy on
  VLSP 2010, 96.77\% of accuracy on VLSP 2013, 94.07\% of $F_1 $ score on NER
  2016, and 90.31\% of $F_1$ score on NER 2018.
\item We release code as open source to facilitate adoption and
  further research, including pre-trained models viBERT and vELECTRA.
\end{itemize}

The remainder of this paper is structured as
follows. Section~\ref{sec:models} presents the methods used in the
current work. Section~\ref{sec:experiments} describes the experimental
results.  Finally, Section~\ref{sec:conclusion} concludes the papers
and outlines some directions for future work.

\section{Models}
\label{sec:models}

\subsection{BERT Embeddings}

\subsubsection{BERT}
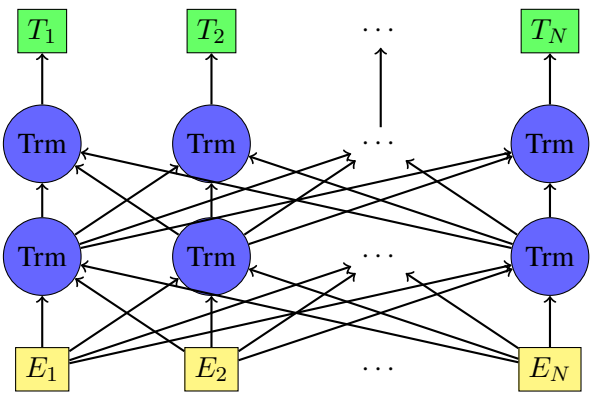
\begin{figure}[t]
  \centering
  \begin{tikzpicture}[x=2.25cm,y=1.5cm]
    \tikzstyle{every node} = [rectangle, draw, fill=gray!30]
    \node[fill=green!60] (a) at (0,4) {$T_1$};
    \node[fill=green!60] (b) at (1,4) {$T_2$};
    \node[fill=none,draw=none] (c) at (2,4) {$\cdots$};
    \node[fill=green!60] (d) at (3,4) {$T_N$};
    \node[circle,fill=blue!60] (xa) at (0,3) {Trm};
    \node[circle,fill=blue!60] (xb) at (1,3) {Trm};
    \node[draw=none,fill=none] (xc) at (2,3) {$\cdots$};
    \node[circle,fill=blue!60] (xd) at (3,3) {Trm};
    \node[circle,fill=blue!60] (ya) at (0,2) {Trm};
    \node[circle,fill=blue!60] (yb) at (1,2) {Trm};
    \node[draw=none,fill=none] (yc) at (2,2) {$\cdots$};
    \node[circle,fill=blue!60] (yd) at (3,2) {Trm};
    \node[draw,fill=yellow!60] (ea) at (0,1) {$E_1$};
    \node[draw,fill=yellow!60] (eb) at (1,1) {$E_2$};
    \node[draw=none,fill=none] (ec) at (2,1) {$\cdots$};
    \node[draw,fill=yellow!60] (ed) at (3,1) {$E_N$};
    \foreach \from/\to in {xa/a, xb/b, xc/c, xd/d,
      ya/xa, ya/xb, ya/xc, ya/xd,
      yb/xa, yb/xb, yb/xc, yb/xd,
      yd/xa, yd/xb, yd/xc, yd/xd,
      ea/ya, ea/yb, ea/yc, ea/yd,
      eb/ya, eb/yb, eb/yc, eb/yd,
      ed/ya, ed/yb, ed/yc, ed/yd}
      \draw[->,thick] (\from) -- (\to);
  \end{tikzpicture}
  \caption{The basic structure of BERT}
  \label{fig:bert}
\end{figure}

The basic structure of BERT~\cite{Devlin:2019} (\textit{Bidirectional
  Encoder Representations from Transformers}) is summarized on
Figure~\ref{fig:bert} where Trm are transformation and $E_k$ are
embeddings of the $k$-th token.

In essence, BERT's model architecture is a multilayer bidirectional
Transformer encoder based on the original implementation described
in~\cite{Vaswani:2017}. In this model, each input token of a sentence
is represented by a sum of the corresponding token embedding, its
segment embedding and its position embedding. The WordPiece embeddings
are used; split word pieces are denoted by \#\#. In our experiments,
we use learned positional embedding with supported sequence lengths up
to 256 tokens.

The BERT model trains a deep bidirectional representation by masking
some percentage of the input tokens at random and then predicting only
those masked tokens. The final hidden vectors corresponding to the
mask tokens are fed into an output softmax over the vocabulary. We use
the whole word masking approach in this work. The masked language
model objective is a cross-entropy loss on predicting the masked
tokens. BERT uniformly selects 15\% of the input tokens for
masking. Of the selected tokens, 80\% are replaced  with [MASK], 10\%
are left unchanged, and 10\% are replaced by a randomly selected
vocabulary token. 

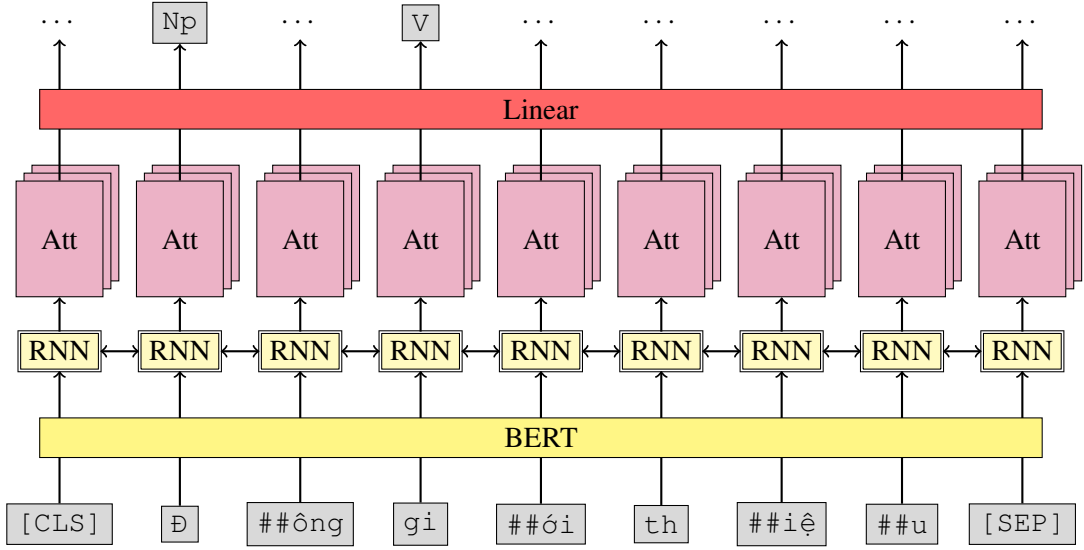
\begin{figure*}
  \centering
  \begin{tikzpicture}[x=1.6cm,y=1.15cm]
    \tikzstyle{every node} = [rectangle,draw,fill=gray!30]
    \foreach \pos in {0,...,8} {
      \node[fill=none,draw=none] (a\pos) at (1*\pos,5) {$\cdots$};
    }
    \node (a1) at (1,5) {\texttt{Np}};
    \node (a3) at (3,5) {\texttt{V}};
    \foreach \pos in {0,...,8} {
      \matt{m\pos}{\pos}{2.5}{Att};
    }
    \foreach \pos in {0,...,8} {
      \draw[->,thick] (m\pos) -- (a\pos);
    }
    \node (c0) at (0,4) {};
    \node (c8) at (8,4) {};
    \node[fill=red!60,align=center,fit={(c0) (c8)}] {};
    \node[fill=none,draw=none] at (4,4) {Linear};
    
    \foreach \pos in {0,...,8} {
      \node[double,fill=yellow!30] (r\pos) at (\pos,1.2) {RNN};
    }
    \foreach \pos in {0,...,7} {
      \draw[<->,thick] (r\pos) -- (r\the\numexpr\pos+1\relax);
    }
   
    \node (t0) at (0,-0.8) {\texttt{[CLS]}};
    \node (t1) at (1,-0.8) {\texttt{Đ}};
    \node (t2) at (2,-0.8) {\texttt{\#\#ông}};
    \node (t3) at (3,-0.8) {\texttt{gi}};
    \node (t4) at (4,-0.8) {\texttt{\#\#ới}};
    \node (t5) at (5,-0.8) {\texttt{th}};
    \node (t6) at (6,-0.8) {\texttt{\#\#iệ}};
    \node (t7) at (7,-0.8) {\texttt{\#\#u}};
    \node (t8) at (8,-0.8) {\texttt{[SEP]}};
    \foreach \pos in {0,...,8} {
      \draw[->,thick] (t\pos) -- (r\pos);
      \draw[->,thick] (r\pos) -- (m\pos);
    }
    \node (B0) at (0,0.2) {};
    \node (B8) at (8,0.2) {};
    \node[fill=yellow!60,align=center,fit={(B0) (B8)}] {};
    \node[fill=none,draw=none] at (4,0.2) {BERT};
  \end{tikzpicture}
  \caption{Our proposed end-to-end architecture}
  \label{fig:arch}
\end{figure*}

In our experiment, we start with the open-source mBERT
package\footnote{\url{https://github.com/google-research/bert/blob/master/multilingual.md}}. We
keep the standard hyper-parameters of 12 layers, 768 hidden units, and
12 heads. The model is optimized with Adam~\cite{Kingma:2015} using
the following parameters: $\beta_1 = 0.9, \beta_2 = 0.999$, $\epsilon
= 1e-6$ and $L_2$ weight decay of $0.01$. 

The output of BERT is computed as follows~\cite{Peters:2018}:
\begin{equation*}
  B_k = \gamma \left( w_0 E_k + \sum_{k=1}^m w_i h_{ki} \right),
\end{equation*}
where
\begin{itemize}
\item $B_k$ is the BERT output of $k$-th token;
\item $E_k$ is the embedding of $k$-th token;
\item $m$ is the number of hidden layers of BERT;
\item $h_{ki}$ is the $i$-th hidden state of of $k$-th token;
\item $\gamma, w_0, w_1,\dots,w_m$ are trainable parameters.
\end{itemize}

\subsubsection{Proposed Architecture}

Our proposed architecture contains five main layers as follows:
\begin{enumerate}
\item The input layer encodes a sequence of tokens which are
  substrings of the input sentence, including ignored indices, padding
  and separators;
\item A BERT layer;
\item A bidirectional RNN layer with either LSTM or GRU units;
\item An attention layer;
\item A linear layer;
\end{enumerate}

A schematic view of our model architecture is shown in
Figure~\ref{fig:arch}.

\subsection{ELECTRA}

\begin{figure*}
  \centering
  \begin{tikzpicture}[x=1.75cm,y=0.8cm]
    \tikzstyle{every node} = [fill=gray!30,text width=1.2cm]
    \node (m1) at (0,4) {phi};
    \node (m2) at (0,3) {công};
    \node (m3) at (0,2) {điều};
    \node (m4) at (0,1) {khiển};
    \node (m5) at (0,0) {máy};
    \node (m6) at (0,-1) {bay};

    \node (n1) at (1,4) {phi};
    \node[fill=green!60] (n2) at (1,3) {MASK};
    \node (n3) at (1,2) {điều};
    \node (n4) at (1,1) {khiển};
    \node[fill=green!60] (n5) at (1,0) {MASK};
    \node (n6) at (1,-1) {bay};

    \node (p1) at (4,4) {phi};
    \node[fill=blue!60] (p2) at (4,3) {công};
    \node (p3) at (4,2) {điều};
    \node (p4) at (4,1) {khiển};
    \node[fill=red!60] (p5) at (4,0) {sân};
    \node (p6) at (4,-1) {bay};

    \foreach \from/\to in {
      m1/n1, m2/n2, m3/n3, m4/n4, m5/n5, m6/n6, n1/p1, n2/p2, n3/p3,
      n4/p4, n5/p5, n6/p6} 
    \draw[->,thick] (\from) -- (\to);
    
    \draw[draw=black,fill=white] (1.75,4.5) rectangle ++(1.5,-6);
    \node[fill=none] (g) at (2.25,2)
      {\textbf{\begin{tabular}{c}Generator\\(BERT)\end{tabular}}};

    \node[fill=none] (q1) at (7,4) {original};
    \node[fill=blue!60] (q2) at (7,3) {original};
    \node[fill=none] (q3) at (7,2) {original};
    \node[fill=none] (q4) at (7,1) {original};
    \node[fill=red!60] (q5) at (7,0) {replaced};
    \node[fill=none] (q6) at (7,-1) {original};

    \foreach \from/\to in {
      p1/q1, p2/q2, p3/q3, p4/q4, p5/q5, p6/q6} 
    \draw[->,thick] (\from) -- (\to);
    
    \draw[draw=black,fill=white] (4.75,4.5) rectangle ++(1.5,-6);
    \node[fill=none] (d) at (5.1,2)
      {\textbf{\begin{tabular}{c}Discriminator\\(ELECTRA)\end{tabular}}};
    
  \end{tikzpicture}
  \caption{An overview of replaced token detection by the ELECTRA
    model on a sample drawn from vELECTRA}
  \label{fig:electra}
\end{figure*}
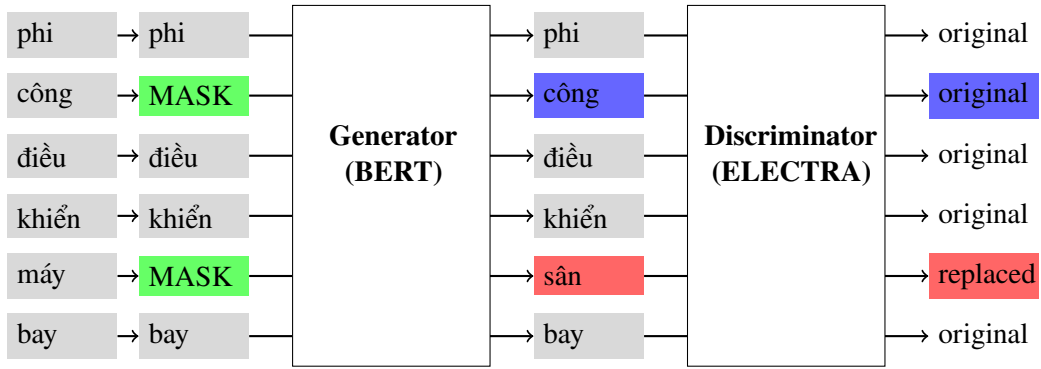

ELECTRA~\cite{Clark:2020} is currently the latest development of BERT-based model
where a more sample-efficient pre-training method is used. This method
is  called replaced token detection. In this method, two neural
networks, a generator $G$ and a discriminator $D$,  are trained
simultaneously. Each one consists of a Transformer network (an
encoder) that maps a sequence of input tokens $\vec x = [x_1,
x_2,\dots,x_n]$ into a sequence of contextualized vectors $h(\vec x) =
[h_1, h_2,\dots, h_n]$. For a given position $t$ where $x_t$ is the
masked token, the generator outputs a probability for generating a
particular token $x_t$ with a softmax distribution:
\begin{equation*}
  p_G(x_t|\vec x) = \frac{\exp(x_t^\top h_G(\vec x)_t) }{\sum_{u}
    \exp(u_t^\top h_G(\vec x)_t)}. 
\end{equation*}
For a given position $t$, the discriminator predicts whether the
token $x_t$ is ``real'', i.e., that it comes from the data rather than the generator distribution, with a
sigmoid function:
\begin{equation*}
D(\vec x, t) = \sigma \left (w^\top h_D(\vec x)_t \right )
\end{equation*}

An overview of the replaced token detection in the ELECTRA model is
shown in Figure~\ref{fig:electra}. The generator is a BERT model which
is trained jointly with the discriminator. The Vietnamese example is a
real one which is sampled from our training corpus.

\section{Experiments}
\label{sec:experiments}

\subsection{Experimental Settings}

\subsubsection{Model Training}
To train the proposed models, we use a CPU (Intel Xeon E5-2699 v4
@2.20GHz) and a GPU (NVIDIA GeForce GTX 1080 Ti 11G). The
hyper-parameters that we chose are as follows: maximum sequence length
is 256, BERT learning rate is $2E-05$, learning rate is $1E-3$, number
of epochs is 100, batch size is 16, use apex and BERT weight decay is
set to 0, the Adam rate is $1E-08$. The configuration of our model is
as follows: number of RNN hidden units is 256, one RNN layer,
attention hidden dimension is 64, number of attention heads is 3 and a
dropout rate of 0.5.

To build the pre-training language model, it is very important to have
a good and big dataset. This dataset was collected from online
newspapers\footnote{vnexpress.net, dantri.com.vn, baomoi.com,
  zingnews.vn, vitalk.vn, etc.} in Vietnamese. To clean the data, we
perform the following pre-processing steps:  
\begin{itemize}
    \item Remove duplicated news
    \item Only accept valid letters in Vietnamese
    \item Remove too short sentences (less than 4 words)
\end{itemize}

We obtained approximately 10GB of texts after collection. This dataset was
 used to further pre-train the mBERT to build our viBERT which better
 represents Vietnamese texts. About the vocab, we removed insufficient
 vocab from mBERT because its vocab contains ones for other
 languages. This was done by keeping only vocabs existed in the
 dataset. 
 
 In pre-training vELECTRA, we collect more data from two sources:
 \begin{itemize}
 \item NewsCorpus: 27.4
   GB\footnote{\url{https://github.com/binhvq/news-corpus}}
 \item OscarCorpus: 31.0
   GB\footnote{\url{https://traces1.inria.fr/oscar/}}
 \end{itemize}

Totally, with more than 60GB of texts, we start training different
versions of vELECTRA. It is worth noting that pre-training viBERT
is much slower than pre-training vELECTRA. For this reason, we
pre-trained viBERT on the 10GB corpus rather than on the large 60GB
corpus. 

\begin{table*}
  \centering
  \begin{tabular}{|l | l | c | c | c | c | c| c | }
    \hline
  \textbf{No.}&\multicolumn{3}{c|}{\textbf{VLSP 2010}}&\multicolumn{4}{c|}{\textbf{VLSP 2013}}\\ \hline

\multicolumn{8}{|l|}{\textbf{Existing models}} \\ \hline
1. & \multicolumn{2}{|l|}{MEM~\cite{Le:2010}}& 93.4 & \multicolumn{3}{|l|}{RDRPOSTagger~\cite{Nguyen:2014}} & 95.1 \\ \hline
2. &  \multicolumn{3}{c}{} & \multicolumn{3}{|l|}{BiLSTM-CNN-CRF~\cite{Ma:2016}}  & 95.4 \\ \hline
3. &  \multicolumn{3}{c}{} & \multicolumn{3}{|l|}{VnCoreNLP-POS~\cite{Nguyen:2017}}  & 95.9 \\ \hline
4. &  \multicolumn{3}{c}{} & \multicolumn{3}{|l|}{jointWPD~\cite{Nguyen:2019}}  & 96.0 \\ \hline
5. &  \multicolumn{3}{c}{} & \multicolumn{3}{|l|}{PhoBERT\_base~\cite{Nguyen:2020}}  & 96.7 \\ \hline

\hline
\multicolumn{8}{|l|}{\textbf{Proposed models}} \\ \hline
  & \textbf{Model Name} & \textbf{mBERT} & \textbf{viBERT} & \textbf{vELEC} & \textbf{mBERT} &	\textbf{viBERT} &  \textbf{vELEC}\\
  \hline
1.&+Fine-Tune & 94.34 & 	95.07 & 95.35 &	96.35 & 	96.60 & 96.62 \\ \hline
2.&+BiLSTM& 94.34	 & 95.12 & 95.32 & 	96.38 & 	96.63 & \textbf{96.77} \\ \hline
3.&+BiGRU& 94.37 &  95.13 &  	95.37 & 96.45 &  	96.68& 96.73\\	 \hline
4.&+BiLSTM\_Attn& 94.37 & 95.12 & \textbf{95.40} & 96.36 & 96.61	 & 96.61 \\\hline
5.&+BiGRU\_Attn& 94.41 &  95.13 &  95.35 & 96.33 &  96.56 & 96.55 \\ \hline
 \end{tabular}
  \caption{Performance of our proposed models on the POS tagging task}
  \label{tab:result-POS}
\end{table*}

\subsubsection{Testing and evaluation methods}
In performing experiments, for datasets without development sets, we
randomly selected 10\% for fine-tuning the best parameters.  

To evaluate the effectiveness of the models, we use the  commonly-used
metrics which are proposed by the organizers of VLSP. Specifically, we
measure the accuracy score on the POS tagging task which is calculated
as follows: 
\begin{equation*}
  Acc = \frac{\#of\_words\_correcly\_tagged}{\#of\_words\_in\_the\_test\_set}
\end{equation*}

and the $F_1$ score on the NER task using the following equations: 
\begin{equation*}
    F_1 = 2*\frac{Pre*Rec}{Pre+Rec}
\end{equation*}
where \textit{Pre} and \textit{Rec} are determined as follows:
\begin{equation*}
    Pre = \frac{NE\_true}{NE\_sys}
\end{equation*}

\begin{equation*}
    Rec = \frac{NE\_true}{NE\_ref}
\end{equation*}
                   
where \textit{NE\_ref} is the number of NEs in gold data,
\textit{NE\_sys} is the number of NEs in recognizing system, and
\textit{NE\_true} is the number of NEs which is correctly recognized
by the system. 

\subsection{Experimental Results}

\subsubsection{On the PoS Tagging Task}

Table~\ref{tab:result-POS} shows experimental results using different
proposed architectures on the top of mBERT and
viBERT and vELECTRA on two  benchmark datasets from
the campaign VLSP 2010 and VLSP 2013.  

As can be seen that, with further pre-training techniques on a
Vietnamese dataset, we could significantly improve the performance of
the model. On the dataset of VLSP 2010, both viBERT and
vELECTRA significantly improved the performance by about 1\%
in the $F_1$ scores.  On the dataset of VLSP 2013, these two models
slightly improved the  performance.  

From the table, we can also see the performance of different
architectures including fine-tuning, BiLSTM, biGRU, and their
combination with attention mechanisms. Fine-tuning mBERT with
linear functions in several epochs could produce nearly state-of-the-art
results. It is also shown that building different architectures on top
slightly improve the performance of all mBERT,
viBERT and vELECTRA models. On the VLSP 2010, we got
the accuracy of 95.40\% using biLSTM with attention on top of
vELECTRA. On the VLSP 2013 dataset, we got 96.77\% in the
accuracy scores using only biLSTM on top of vELECTRA.    

In comparison to previous work, our proposed model - vELECTRA
- outperformed previous ones. It achieved from 1\% to 2\% higher than
existing work using different innovation in deep learning such as CNN,
LSTM, and joint learning techniques. Moreover,  vELECTRA also
gained a slightly better than PhoBERT\_base, the same
pre-training language model released so far, by nearly 0.1\% in the
accuracy score.

\begin{table*}
  \centering
  \begin{tabular}{|l | l | c | c | c | c |c | c | }
    \hline
  \textbf{No.}& \multicolumn{4}{c|}{\textbf{VLSP 2016}}&\multicolumn{3}{c|}{\textbf{VLSP 2018}}\\ \hline
\multicolumn{8}{|l|}{\textbf{Existing models}} \\ \hline

 1. & \multicolumn{3}{|l|}{TRE+BI~\cite{Le:2016}} & 87.98 &  \multicolumn{2}{|l|}{VietNER} & 76.63 \\ \hline
 
2. & \multicolumn{3}{|l|}{BiLSTM\_CNN\_CRF~\cite{Pham:2017a}} & 88.59 &  \multicolumn{2}{|l|}{ZA-NER} & 74.70 \\ \hline
 3. & \multicolumn{3}{|l|}{BiLSTM~\cite{Pham:2017c}} & 92.02 &   \multicolumn{3}{|l|}{} \\ \hline
  4. & \multicolumn{3}{|l|}{NNVLP~\cite{Pham:2017b}} & 92.91 &   \multicolumn{3}{|l|}{} \\ \hline

5. & \multicolumn{3}{|l|}{VnCoreNLP-NER~\cite{Vu:2018}} & 88.6 & \multicolumn{3}{|l|}{} \\ \hline
6. & \multicolumn{3}{|l|}{VNER~\cite{Nguyen:2019}} & 89.6 &  \multicolumn{3}{|l|}{}\\ \hline
7. & \multicolumn{3}{|l|}{ETNLP~\cite{Vu:2019}} & 91.1 & \multicolumn{3}{|l|}{}  \\ \hline
8. & \multicolumn{3}{|l|}{PhoBERT\_base~\cite{Nguyen:2020}} & 93.6 & \multicolumn{3}{|l|}{}  \\ \hline
 
\multicolumn{8}{|l|}{\textbf{Proposed models}} \\ \hline
  &\textbf{Model Name}& \textbf{mBERT} & \textbf{viBERT} & \textbf{VELEC} & \textbf{mBERT} &	\textbf{viBERT} &  \textbf{VELEC} \\
  \hline
1.&+Fine-Tune & 91.28 & 	92.84 & 94.00 & 	86.86 & 	88.04 & 89.79 \\ \hline
2.&+BiLSTM& 91.03 & 	93.00 & 93.70 &	86.62 & 	88.68 & 89.92 \\ \hline
3.&+BiGRU& 91.52 &  93.44 & 93.93 & 86.72 &  88.98 & \textbf{90.31}\\	 \hline
4.&+BiLSTM\_Attn& 91.23 & 	92.97 & \textbf{94.07} &	87.12& 	89.12 & 90.26 \\\hline
5.&+BiGRU\_Attn& 90.91 &  93.32 & 93.27 &	86.33 &   88.59 &89.94 \\ \hline
 \end{tabular}
  \caption{Performance of our proposed models on the NER task. ZA-NER~\cite{Luong:2018}
    is the best system of VLSP 2018~\cite{Nguyen:2018}. VietNER is
    from~\cite{NguyenKA:2019}} 
  \label{tab:result-NER}
\end{table*}

\subsubsection{On the NER Task}

Table~\ref{tab:result-NER} shows experimental results using different
proposed architectures on the top of mBERT,  viBERT
and vELECTRA on two  benchmark datasets from the campaign
VLSP 2016 and VLSP 2018.  

These results once again gave a strong evidence to the above statement
that further training mBERT on a small raw dataset could
significantly improve the performance of transformation-based language
models on downstream tasks. Training vELECTRA from scratch on
a big Vietnamese dataset could further enhance the performance. On two
datasets, vELECTRA improve the $F_1$ score by from 1\% to 3\% in
comparison to viBERT and mBERT.  

Looking at the performance of different architectures on top of these
pre-trained models, we acknowledged that biLSTM with attention once a
gain yielded the SOTA result on VLSP 2016 dataset. On VLSP 2018
dataset, the architecture of biGRU yielded the best performance at
90.31\% in the $F_1$ score. 

Comparing to previous work, the best proposed model outperformed all
work by a large margin on both datasets.

\begin{figure*}
  \centering
  \begin{tikzpicture}
    \begin{axis}[
      height=5.8cm,
      width=\textwidth,
      ybar,
      bar width=10,
      enlarge y limits=0.25,
      legend style={at={(0.5,0.85)},anchor=south,legend columns=-1},
      xlabel={\textit{}},
      ylabel={\textit{milliseconds per sentence}},
      symbolic x coords={+Fine-Tune,+biLSTM,+biGRU,+biLSTM-Att,+biGRU-Att},
      xtick=data,
      nodes near coords,
      every node near coord/.append style={font=\tiny,rotate=90,anchor=east},
      nodes near coords={\pgfmathprintnumber[precision=3]{\pgfplotspointmeta}},
      nodes near coords align={vertical},
      ]
      \addplot coordinates {(+Fine-Tune,1.8) (+biLSTM,2.8) (+biGRU,3.1)  (+biLSTM-Att, 2.9) (+biGRU-Att, 3.3)};
      \addplot coordinates {(+Fine-Tune,2.9) (+biLSTM,2.8) (+biGRU,2.7)  (+biLSTM-Att, 2.9) (+biGRU-Att, 2.9)};
      \addplot coordinates {(+Fine-Tune,1.6) (+biLSTM,2.4) (+biGRU,2.6)  (+biLSTM-Att, 1.8) (+biGRU-Att, 2.4)};
      \legend{mBERT, viBERT, vELECTRA}
    \end{axis}
  \end{tikzpicture}
  \caption{Decoding time on PoS task -- VLSP 2013}
  \label{fig:vlsp2013}  
\end{figure*}
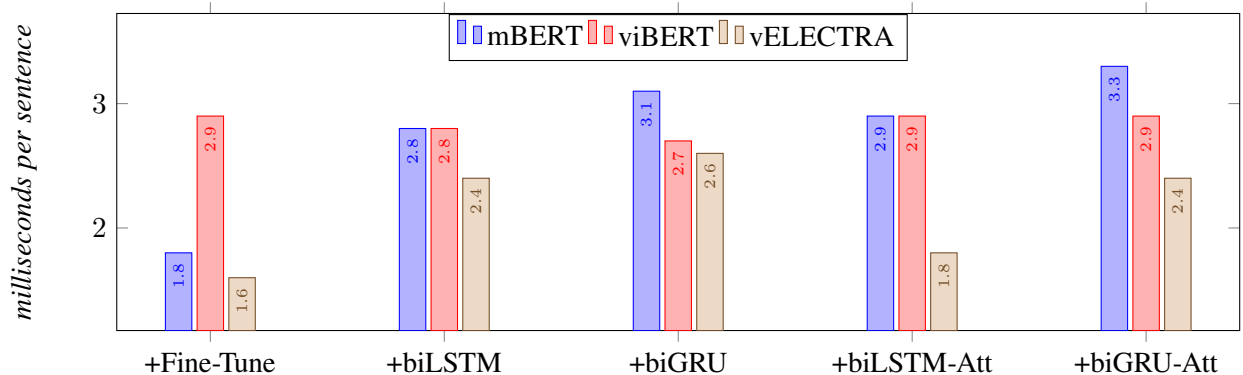

\subsection{Decoding Time}

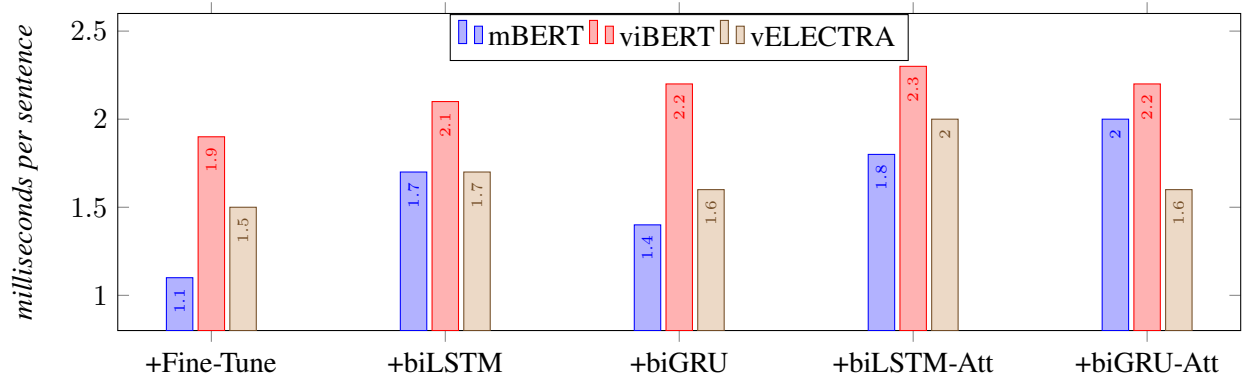
\begin{figure*}
  \centering
  \begin{tikzpicture}
    \begin{axis}[
      height=5.8cm,
      width=\textwidth,
      ybar,
      bar width=10,
      enlarge y limits=0.25,
      legend style={at={(0.5,0.85)},anchor=south,legend columns=-1},
      xlabel={\textit{}},
      ylabel={\textit{milliseconds per sentence}},
      symbolic x coords={+Fine-Tune,+biLSTM,+biGRU,+biLSTM-Att,+biGRU-Att},
      xtick=data,
      nodes near coords,
      every node near coord/.append style={font=\tiny,rotate=90,anchor=east},
      nodes near coords={\pgfmathprintnumber[precision=3]{\pgfplotspointmeta}},
      nodes near coords align={vertical},
      ]
      \addplot coordinates {(+Fine-Tune,1.1) (+biLSTM,1.7) (+biGRU,1.4)  (+biLSTM-Att, 1.8) (+biGRU-Att, 2.0)};
      \addplot coordinates {(+Fine-Tune,1.9) (+biLSTM,2.1) (+biGRU,2.2)  (+biLSTM-Att, 2.3) (+biGRU-Att, 2.2)};
      \addplot coordinates {(+Fine-Tune,1.5) (+biLSTM,1.7) (+biGRU,1.6)  (+biLSTM-Att, 2.0) (+biGRU-Att, 1.6)};
      \legend{mBERT, viBERT, vELECTRA}
    \end{axis}
  \end{tikzpicture}
  \caption{Decoding time on NER task -- VLSP 2016}
  \label{fig:vlsp2016}  
\end{figure*}

Figure \ref{fig:vlsp2013} and \ref{fig:vlsp2016} shows the averaged
decoding time measured on one sentence. According to our statistics,
the averaged length of one sentence in VLSP 2013 and VLSP 2016
datasets are 22.55 and 21.87 words, respectively.  

For the POS tagging task measured on VLSP 2013 dataset, among three
models, the fastest decoding time is of vELECTRA model, followed by
viBERT model, and finally by mBERT model. This statement holds for
four proposed architectures on top of these three models. However, for
the fine-tuning technique, the decoding time of mBERT is faster than
that of viBERT. 

For the NER task measured on the VLSP 2016 dataset, among three models,
the slowest time is of viBERT model with more than 2
milliseconds per sentence. The decoding times on mBERT topped with
simple fine-tuning techniques, or biGRU, or biLSTM-attention is a
little bit faster than on vELECTRA with the same architecture.  

This experiment shows that our proposed models are of practical
use. In fact, they are currently deployed as a core component of our
commercial chatbot engine FPT.AI\footnote{\url{http://fpt.ai/}} which
is serving effectively many customers. More precisely, the FPT.AI
platform has been used by about 70 large enterprises, and of over 27,000
frequent developers, serving more than 30 million end
users.\footnote{These numbers are reported as of August, 2020.}

\section{Conclusion}
\label{sec:conclusion}

This paper presents some new model architectures for sequence tagging
and our experimental results for Vietnamese part-of-speech tagging and
named entity recognition.  Our proposed model vELECTRA outperforms
previous ones. For part-of-speech tagging, it improves about 2\% of
absolute point in comparison with existing work which use different
innovation in deep learning such as CNN, LSTM, or joint learning
techniques.  For named entity recognition, the vELECTRA outperforms
all previous work by a large margin on both VLSP 2016 and VLSP 2018
datasets.

Our code and pre-trained models are published as an open source
project for facilitate adoption and further research in the Vietnamese
language processing community.\footnote{viBERT is available at
  \url{https://github.com/fpt-corp/viBERT} and vELECTRA is available
  at \url{https://github.com/fpt-corp/vELECTRA}.} An online service of
the models for demonstration is also accessible at
\url{https://fpt.ai/nlp/bert/}.  A variant and more advanced version
of this model is currently deployed as a core component of our
commercial chatbot engine FPT.AI  which is serving effectively
millions of end users. In particular, these models are being
fine-tuned to improve task-oriented dialogue in mixed and multiple
domains~\cite{Luong:2019} and dependency parsing~\cite{Le:2015c}.

\section*{Acknowledgement}

We thank three anonymous reviewers for their valuable comments for
improving our manuscript.





\bibliographystyle{acl}
\bibliography{references}

\end{document}